%% file: kronberger.tex
\documentclass[graybox]{svmult}

\usepackage{mathptmx}       
\usepackage{helvet}         
\usepackage{courier}        
\usepackage{makeidx}         
\usepackage{graphicx}        
\usepackage{multicol}        
\usepackage[bottom]{footmisc}
\usepackage[vlined,boxed,linesnumbered]{algorithm2e} 

\usepackage[caption=false]{subfig}

\usepackage{tikz} 
\usetikzlibrary{shapes,arrows, decorations.markings}
\usetikzlibrary{positioning}

\makeindex             

\newcommand{\cupdot}{\mathbin{\mathaccent\cdot\cup}}

\begin{document}

\title*{Cluster Analysis of a Symbolic Regression Search Space}
\author{Gabriel Kronberger and Lukas Kammerer and Bogdan Burlacu and Stephan M. Winkler and Michael Kommenda and Michael Affenzeller}

\authorrunning{Kronberger et al.}

\institute {Gabriel Kronberger$^{1,3}$ \email{gabriel.kronberger@fh-ooe.at}\and Lukas Kammerer$^{1,2,3}$ \and Bogdan Burlacu$^{1,2,3}$ \and Stephan M. Winkler$^{1,2}$ \and Michael Kommenda$^{1,2,3}$ \and Michael Affenzeller$^{1,2}$
    \at
    \inst{1} Heuristic and Evolutionary Algorithms Laboratory (HEAL),
    University of Applied Sciences Upper Austria, Softwarepark 11, 4232 Hagenberg, Austria\\
    \inst{2} Institute for Formal Models and Verification,
    Johannes Kepler University, Altenberger Stra{\ss}e 69, 4040 Linz, Austria \\
    \inst{3} Josef Ressel Center for Symbolic Regression,
    University of Applied Sciences Upper Austria, Softwarepark 11, 4232 Hagenberg, Austria\\
}
\maketitle
\renewcommand{\thefootnote}{}
\footnotetext{\hspace{-0em}
	The final publication is available at \url{https://link.springer.com/chapter/10.1007/978-3-030-04735-1\_5}. 
}
\renewcommand\thefootnote{\arabic{footnote}}

\abstract*{\input{abstract}}

\abstract{\input{abstract}}

\keywords{symbolic regression, grammar enumeration, approximate nearest neighbors, dimensionality reduction, clustering}

\section{Introduction}
\label{sec:introduction}
Knowledge discovery systems such as Genetic Programming (GP) for symbolic regression often have to deal with a very large search space of mathematical expressions, which only grows exponentially larger with the number of input variables.

Genetic programming guides the search via selection and discovers new model structures via the action of crossover and mutation. Population diversity plays an important role in this process, as it affects the algorithm's ability to assemble existing building blocks into increasingly-fit solutions. The relationship between diversity at both the genotypic and phenotypic level has been previously explored \cite{gustafson:2004:IEEE}, leading to a number of important insights:
\begin{itemize}
        \item Strong exploitation of structures occurs in almost all runs
        \item Diversity at the structural level is quickly lost
        \item Encouraging different amounts of diversity can lead to better performance
        \item The interplay between genetic operators induces a neighborhood structure in the search space
        \item Fitness is positively-correlated with fitness-based (phenotypic) diversity and negatively correlated with genotypic diversity
\end{itemize}

In light of the above, we set our goal to investigate GP's ability to explore different areas of the search space by superimposing a neighborhood structure obtained via clustering of symbolic regression models generated via grammar enumeration.

In this contribution we concentrate on the distribution of models in
symbolic regression search spaces. Our motivation for this work is
that we hope to be able to reduce the computational effort which is
required to find well-fitting symbolic regression models by
precomputing a clustering of all symbolic regression models in the
search space. In particular, we aim to precompute a similarity network
or (equivalently) a hierarchical clustering for symbolic regression
models, that is independent from a concrete dataset.

Our research questions in this work are:

\begin{enumerate}
    \item {\em What is the distribution of models in a symbolic
      regression search space?} We are interested in identifying
      clusters of similar models whereby similarity could either be
      determined based on the model outputs (phenotypic) or similarity of the
      evolved expressions (genotypic).
    \item {\em What is the distribution of solutions visited by
      genetic programming?} Here we are interested in
      the systematic search biases of GP. In particular, (i) whether there
      are areas of the search space that are completely ignored by GP,
      and (ii) how the individuals in a GP population are distributed
      in the search space and how this distribution changes from the
      beginning to the end of a GP run.
\end{enumerate}

Our assumptions about the goals of symbolic regression modeling are the
following. We use these assumptions as a guide for our research in
symbolic regression solution methods.
\begin{itemize}
  \item The aim of symbolic regression modeling is primarily to find
    compact expressions that are open for interpretation.
  \item Shallow expressions are preferred over deeply nested expressions.
  \item Models for real-world regression problems are often made up of
    multiple terms which capture independent effects. The independent
    terms can be modeled one after another.
  \item Interesting real-world regression problems often necessitate
    to capture non-linear correlations in the model. A non-linear
    effect is often driven by only one independent
    variable.
  \item Interactions of two or three variables are
    common. Interactions of more than four variables are often not relevant.
  \item The set of potentially relevant variables is much larger than
    the set of actually relevant variables. It is usually not known
    which individual variables or which interactions of variables are
    relevant.
  \item Measurements of input variables as well as target variables are noisy.
\end{itemize}

\section{Methodology}\label{sec:methodology}
In our journey to find answers for the research questions stated above
we enumerate the complete search space (see Section
\ref{sec:grammar-enum}) and evaluate all expressions for fixed input data. To limit the
size of the search space we consider only uni-variate functions and
limit the maximum length of expressions. Additionally, we use a
grammar which constrains the complexity of expressions and which does
not allow numeric parameters (i.e. random constants). This further reduces the search space.

For the analysis of the distribution of expressions in the symbolic
regression search space (\emph{RQ1}), our idea is to create a visual
map of all expressions which hopefully allows us to identify larger
clusters of similar expressions. Thus, we use the set of all evaluated
expressions, identify the set of phenotypically distinct expressions
and determine the phenotypically nearest neighbors for each expression
(see Section \ref{sec:genotypic-similarity}). This allows us to map
expressions to a two-dimensional space while preserving the local
neighborhood structure of the high-dimensional space. The graph of
nearest neighbors also allows us to create a clustering of all
expressions. As we are interested in both -- the map of phenotypically
similar expressions as well as the map of genotypically similar
expressions -- we create a similar map based on a measure for
genotypic similarity (see Section
\ref{sec:phenotypic-similarity}). Ideally, we expect to see similar
cluster structure on both levels, assuming that expressions that are
genotypically similar should also be phenotypically similar and
vice-versa\footnote{We actually found that this assumption is
  wrong. We found that the search space can be split into clusters
  of phenotypically and genotypically similar expressions. However, we
  could not show that phenotypically similar expressions also are
  phenotypically similar and/or vice versa. This is intuitive because two
  highly similar expressions become dissimilar on the phenotypic level
  just by a multiplication with zero. Symmetrically, many different
  expressions can be found which produce the same output.}.

For the analysis of the search bias of GP (\emph{RQ2}), the idea is to
re-use the map and clusters that we generated in the first phase and
analyze whether GP explores the complete map and all clusters. Our
idea is to find the phenotypically most similar expression in the
pre-calculated map for each solution candidate visited by GP. For this
we determine the visitation frequency for the pre-computed clusters
for each generation of GP. We expect that GP visits many different
clusters in the beginning and converges to the clusters with
well-fitting expressions at the end of the run.

\begin{figure}
  \centering
\tikzstyle{block} = [rectangle, draw, text width=3.1cm, text centered, minimum height=1.2cm]
\tikzstyle{line} = [draw, -latex']

\begin{tikzpicture}[node distance=1.7cm, auto]
\node [block, text width = 2.5cm] (GP) {GP};
\node [block, right=1.15cm of GP, text width=2.5cm] (eval) {Evaluated expressions};
\node [block, below of=GP, text width=2.5cm] (exactNN) {Exact nearest neighbors};
\node [below of=eval](pos){};

\node [block, below of=exactNN, text width=2.5cm] (mapGPSols) {Map to embedding\\ and clustering};

\node [below of=mapGPSols](posGP){};
\node [block, left=.05cm of posGP, text width=2.5cm] (clusterFreq) {Cluster visitation frequency};
\node [block, right=.05cm of posGP, text width=2.5cm] (embedding) {GP solutions in embedding};

\tikzset{arrow/.style={decoration={markings,mark=at position 1 with %
    {\arrow[scale=2,>=stealth]{>}}},postaction={decorate}}}

\draw[arrow] (GP) -- (exactNN);
\draw[arrow] (eval) -- (exactNN);
\draw[arrow] (exactNN) -- (mapGPSols);
\draw[arrow] (mapGPSols) -- (clusterFreq);
\draw[arrow] (mapGPSols) -- (embedding);

\end{tikzpicture}
\caption{Overview of the flow of information for tracking which parts of the search space are explored by GP (\emph{RQ2}). GP solution candidates are mapped to the visualization and the clusters by finding the most similar representative.}
\label{fig:methodology-rq2}
\end{figure}
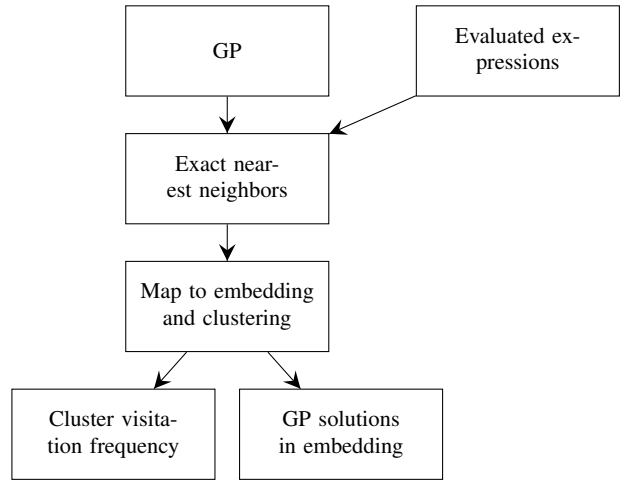

Figure \ref{fig:methodology-rq1} shows an overview of the flow of information for
 search space visualization and clustering. Figure \ref{fig:methodology-rq2} shows how we map GP solution candidates to the pre-computed mapping of the search space and the clusters.

A major challenge in our methodological approach is the sheer size of
the search space. We have handled this challenge using
state-of-the-art algorithms for dimensionality reduction and
clustering which still work for datasets with millions of observations
and hundreds of features (see Section
\ref{sec:clustering-and-viz}). The core idea of these algorithms is
the approximation of nearest neighbors using random projection trees \cite{dasgupta2008random}.

\begin{figure}
\centering
\tikzstyle{block} = [rectangle, draw, text width=3.1cm, text centered, minimum height=1.2cm]
\tikzstyle{line} = [draw, -latex']

\begin{tikzpicture}[node distance=1.7cm, auto]
\node [block] (grammar) {Grammar (Lst. \ref{lst:grammar})};
\node [block, below of=grammar] (generate) {Generate all expressions (Lst. \ref{lst:sentence_generation})};
\node [block, below of=generate] (filterUnique) {Filter unique expressions with hashing};
\node [block, below of=filterUnique] (eval) {Evaluate remaining expressions};

\node [below of=eval](pos){};

\node [block, right=1.15cm of pos] (filterBad) {Filter for $R^2> 0.2$};
\node [block, below of=filterBad] (calcGenotyp) {Calculate genotypic similarity (Sec. \ref{sec:genotypic-similarity})};
\node [below of=calcGenotyp](posRight){};
\node [block, right =.05cm of posRight, text width =2.5cm] (clusterGenotyp) {HDBSCAN (Fig. \ref{fig:genotypic-clustering})};
\node [block, left =.05cm of posRight, text width =2.5cm] (tsne) {t-SNE (Fig. \ref{fig:genotypic-clustering})};

\node [block, left=1.15cm of pos] (approxNN) {Approximate nearest neighbors (Sec. \ref{sec:phenotypic-clustering})};
\node [below of=approxNN](posLeft){};
\node [block, left =.05cm of posLeft, text width =2.5cm] (clusterPheno) {HDBSCAN (Fig. \ref{fig:phenotypic-clustering})};
\node [block, right =.05cm of posLeft, text width =2.5cm] (largeVis) {LargeVis (Fig. \ref{fig:phenotypic-clustering})};

\node [above=.01cm of filterBad]{\large Genotypic};
\node [above=.01cm of approxNN]{\large Phenotypic};

\tikzset{arrow/.style={decoration={markings,mark=at position 1 with %
    {\arrow[scale=2,>=stealth]{>}}},postaction={decorate}}}

\draw[arrow] (grammar) -- (generate);
\draw[arrow] (generate) -- (filterUnique);
\draw[arrow] (filterUnique) -- (eval);
\draw[arrow] (eval) -- (approxNN);
\draw[arrow] (eval) -- (filterBad);

\draw[arrow] (approxNN) -- (clusterPheno);
\draw[arrow] (approxNN) -- (largeVis);

\draw[arrow] (filterBad) -- (calcGenotyp);
\draw[arrow] (calcGenotyp) -- (tsne);
\draw[arrow] (calcGenotyp) -- (clusterGenotyp);

\end{tikzpicture}
\caption{Overview of the flow of information for the search space visualization and clustering (\emph{RQ1}).}
\label{fig:methodology-rq1}
\end{figure}
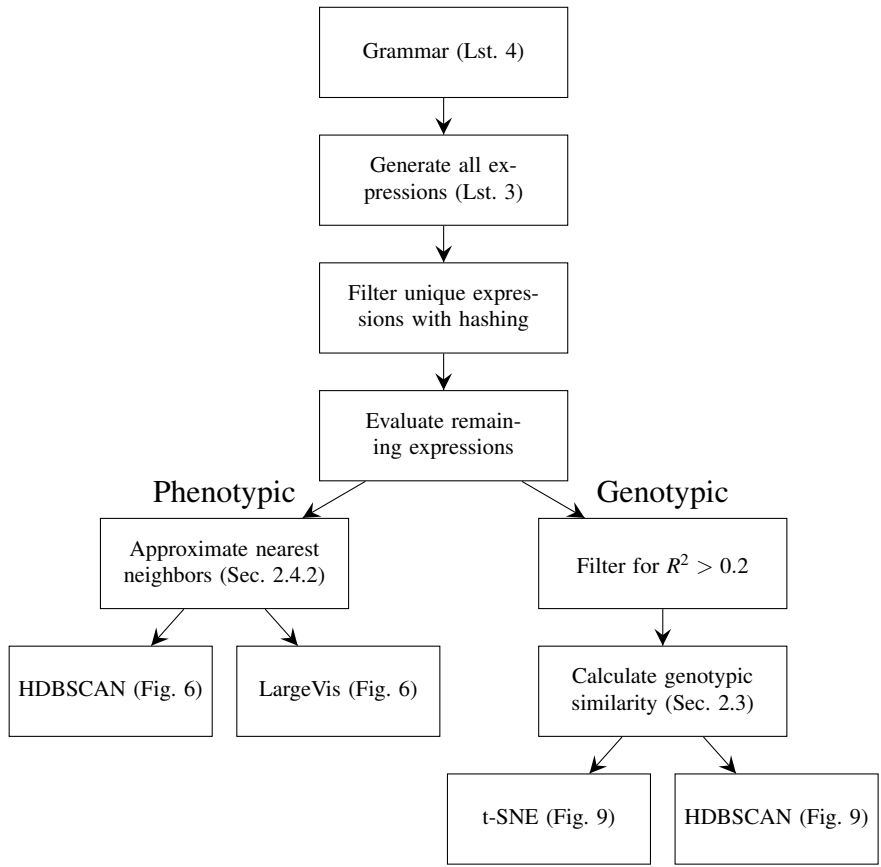

\subsection{Grammar Enumeration}
\label{sec:grammar-enum}
We create symbolic regression models by deriving sentences from a formal grammar for expressions as shown in the listing in Figure \ref{lst:sentence_generation}. By defining a maximum sentence length and omitting numerical constants as a start, a large but finite set of all possible models -- the search space -- can be generated for a problem. These sentences without actual constants values can be seen as a general structure of an actual model (\cite{kommenda2013effects},\cite{worm2013prioritized}).

\begin{figure}
\begin{verbatim}
stack.push(start)

while stack not empty:
   phrase = stack.pop()
   symbol = Fetch nonterminal symbol from phrase
   for each production rule of symbol:
      create new phrase with substituted symbol

      if new phrase is a sentence:
         evaluate(new phrase)
         save(new phrase) 
      else:
         stack.push(new phrase)
\end{verbatim}
\caption{Pseudo-code for generating all sentences of a language defined via a context-free grammar.}
\label{lst:sentence_generation}
\end{figure}

Given a formal grammar many mathematical identities and other phenotypically equal but genotypically different expressions are generated. This includes for example different orders of arguments in commutative operators or different representations of binomial identities.
To keep the search space size manageable, we want to avoid semantic duplicates. Although it is computationally not feasible to fully prevent all semantic duplicates in such a large search space, their number can be largely reduced in two simple steps: First, the grammar is restricted, so that only one representation of relevant mathematical identities can be derived. Second, identities which cannot be prevented in the grammar are identified by hashing: semantic duplicates should have the same hash value.

Expressions derived from the restricted grammar (Figure \ref{lst:grammar}) are sums of terms, which again contain variables and unary functions, such as e.g.~the sine function or the inverse function. The latter can only occur once per term. Also the structures of function arguments are individually restricted.

\begin{figure}
\begin{verbatim}
G(Expr):
Expr   -> Term "+" Expr | Term
Term   -> Factor "*" Term | Factor | "1/(" InvExpr ")"
Factor -> VarFac | ExpFac | LogFac | SinFac
VarFac -> <variable>
ExpFac -> "exp(" SimpleTerm ")"
LogFac -> "log(" SimpleExpr ")"
SinFac -> "sin(" SimpleExpr ")"

SimpleExpr -> SimpleTerm "+" SimpleExpr | SimpleTerm
SimpleTerm -> VarFac "*" SimpleTerm | VarFac

InvExpr -> InvTerm "+" InvExpr | InvTerm
InvTerm -> Factor "*" InvTerm | Factor
\end{verbatim}
\caption{The formal grammar used for grammar enumeration. In the design of the grammar, we want to allow a large set of potentially interesting expressions on the one hand, on the other hand we want to restrict the search space to disallow overly complex expressions as well as many different forms of semantically equal expressions.}
\label{lst:grammar}
\end{figure}

Using a context-free grammar, semantic duplicates such as differently ordered terms cannot be prevented. Therefore, we additionally use a semantic hash function to identify semantically equivalent expressions. For each derived symbolic expression we calculate a hash value symbolically without evaluating the expression. Semantic duplicates are recognized by comparing the hash values of previously derived sentences. In case of a collision, the derived sentence is a likely to be a semantic duplicate and therefore discarded.

The hashing function calculates the hash value recursively from a syntax tree. Each terminal symbol in the tree is assigned to a constant hash value. To cover commutativity, binary operators like multiplication or addition are flattened to n-ary operators and their arguments are ordered.

\subsection{Phenotypic Similarity}
\label{sec:phenotypic-similarity}

For the phenotypic similarity we use Pearson's correlation coefficient
of the model outputs. This allows us to determine the output similarity regardless of the offset and scale of the function values
\cite{keijzer2003improving}. When we evaluate the expressions it is
necessary to assume a range of valid input values. We use 100 points
distributed on a grid in the range $(-5.0 .. 5.0)$. All output vectors
are scaled to zero mean and unit variance; undefined output values and
infinity values are replaced by the average output. This preprocessing allows us to
use cosine-similarity for the clustering and visualization which is
supported by many implementations for approximate nearest neighbors and equivalent to Pearson's
correlation coefficient for zero-mean vectors.

\subsection{Genotypic Similarity}
\label{sec:genotypic-similarity}

We define the genotypic similarity between two tree-based solution candidates using the S{\o}rensen-Dice index
\begin{equation}
        \textrm{GenotypicSimilarity}(T_1, T_2) = \frac{2 \cdot |M|}{|T_1| + |T_2|}\label{eq:genotypic-similarity}
\end{equation}
Here, $M$ is the bottom-up mapping between $T_1$ and $T_2$ calculated using the algorithm described in \cite{Valiente01anefficient}. We describe below the main steps of the algorithm:
\begin{enumerate}
        \item Build a forest $F = T_1 \cupdot T_2$ consisting of the disjoint union between the two trees
        \item Map $F$ to a directed acyclic graph $G$. Two nodes in $F$ are mapped to the same vertex in $G$ if they are at the same height in the tree and their children are mapped to the same sequence of vertices in $G$. The bottom-up traversal ensures that nodes are mapped before their parents, leading to $O(|T_1| + |T_2)|$ build time for $G$.
        \item Use the map $K : F \to G$ obtained in the previous step to build the final mapping $M : T_1 \to T_2$. This step iterates over the nodes of $T_1$ in level order and uses $K$ to determine which nodes correspond to the same vertices in $G$. Level-order iteration guarantees that every largest unmapped subtree of $T_1$ will be mapped to an isomorphic subtree of $T_2$.
\end{enumerate}
The algorithm has a runtime complexity linear in the size of the trees regardless whether the trees are ordered or unordered.

\subsection{Clustering and visualization}
\label{sec:clustering-and-viz}
One of the challenges in visualizing the search space using
phenotypic or genotypic similarity measures is to find a mapping of expressions to a two-dimensional space which preserves pairwise similarities as well as possible.

We use the t-SNE algorithm \cite{maaten2008visualizing} with the distance matrices which are calculated on the basis of the previously described phenotypic or genotypic similarity measures.

The main idea behind t-SNE is to map a high-dimensional space $X$ to a low-dimensional space $Y$ where the distribution of pairwise similarities is preserved as much as possible. Similarity between data points $x_i,x_j \in X$ is defined as the probability that $x_i$ would pick $x_j$ as its neighbor. A similar probability distribution is found in $Y$ by minimizing the Kullback-Leibler divergence between the two distributions using gradient descent.
While t-SNE does not preserve distances, the visualization can potentially provide new insight into the structure of the search space for symbolic regression.

In the following we describe the visualization and clustering procedure in more detail.

\subsubsection{Clustering and visualization based on genotypic similarity}
\begin{figure}
        \includegraphics[width=\textwidth]{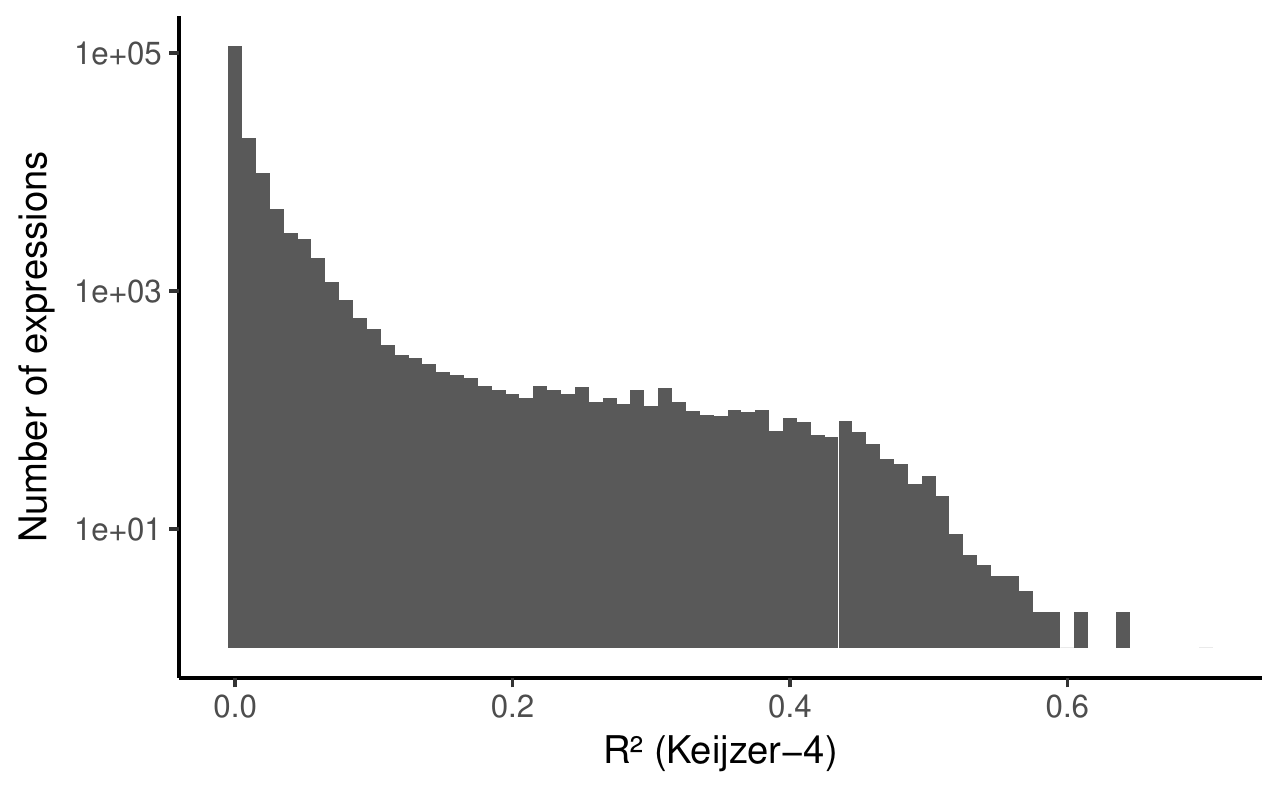}
        \caption{Distribution of $R^2$ values for the Keijzer-4 function}
        \label{fig:quality-distribution}
\end{figure}

The base set of $\approx 1.6 \cdot 10^5$ unique expressions obtained via grammar enumeration is unfeasibly big for the calculation of the full similarity matrix. Therefore, we further reduce this set to a feasible quantity by filtering expressions based on their $R^2$ value in relation to the Keijzer-4 function \cite{keijzer2003improving}. Figure \ref{fig:quality-distribution} shows the distribution of $R^2$ values over the full set of all unique expressions. For the genotypic mapping we took only expressions with $R^2 > 0.2$.

We used the HDBSCAN algorithm \cite{Campello2013} for clustering. We tried two approaches: (i) clustering based directly on the pre-computed similarity matrix, and (ii) clustering in the mapped 2-d space. Both approaches produced similar results.

\subsubsection{Clustering and visualization based on phenotypic similarity}
\label{sec:phenotypic-clustering}
For the mapping based on phenotypic similarity we decided to use the
complete set of the unique expressions and applied the \emph{LargeVis}
implementation \cite{Tang2016} to produce the visualization. LargeVis
relies on approximate nearest neighbor algorithms to make
visualization of large-scale and high-dimensional datasets
feasible. For our analysis we used the R library for
LargeVis\footnote{\url{https://github.com/elbamos/largeVis}}.
LargeVis implements a variant of t-SNE in which the exact
determination of nearest neighbors is replaced by the approximate
nearest neighbors list. This has only linear runtime
complexity in the number of data points. As a consequence, the asymptotic runtime of clustering and
the embedding becomes linear in the number of data points.

The algorithm works in three steps. First, a the lists of approximate
nearest neighbors for each data points are determined using random projection trees \cite{dasgupta2008random}. In the second step, a sparse weighted edge matrix
is calculated which encodes the nearest neighbor graph. Finally, the
approximate nearest neighbor lists and the edge matrix can be used for
t-SNE. LargeVis provides a variant of the HDBSCAN algorithm
\cite{McInnes2017icdmw,McInnes2017soft} which uses the approximate
nearest neighbor list.

\subsection{Mapping GP solution candidates}
Based on the results of the phenotypic clustering we study how GP
explores the search space. For this we add a step in the GP algorithm
after all individuals in the population have been evaluated. In this
step, we identify the expression or cluster of expressions in the
enumerated search space for each evaluated solution candidate. With
this we are able to calculate a visitation density for the enumerated
search space. 

We expect that in the early stages of evolution GP explores many
different areas of the search space, whereby over time GP should
converge to the area of the search space which contains expressions
which are most similar to the target function.

\section{Results}
\label{sec:results}
In the following sections we first present our results of the clustering and visualization based on the phenotypic similarity and compare with the results of clustering and visualization based on the genotypic similarity. Then we present the results of our analysis of cluster qualities for five uni-variate functions. Finally, we present the results of the GP visitation frequency analysis.

\subsection{Phenotypic mapping}
\begin{figure}
  \includegraphics[width=\textwidth]{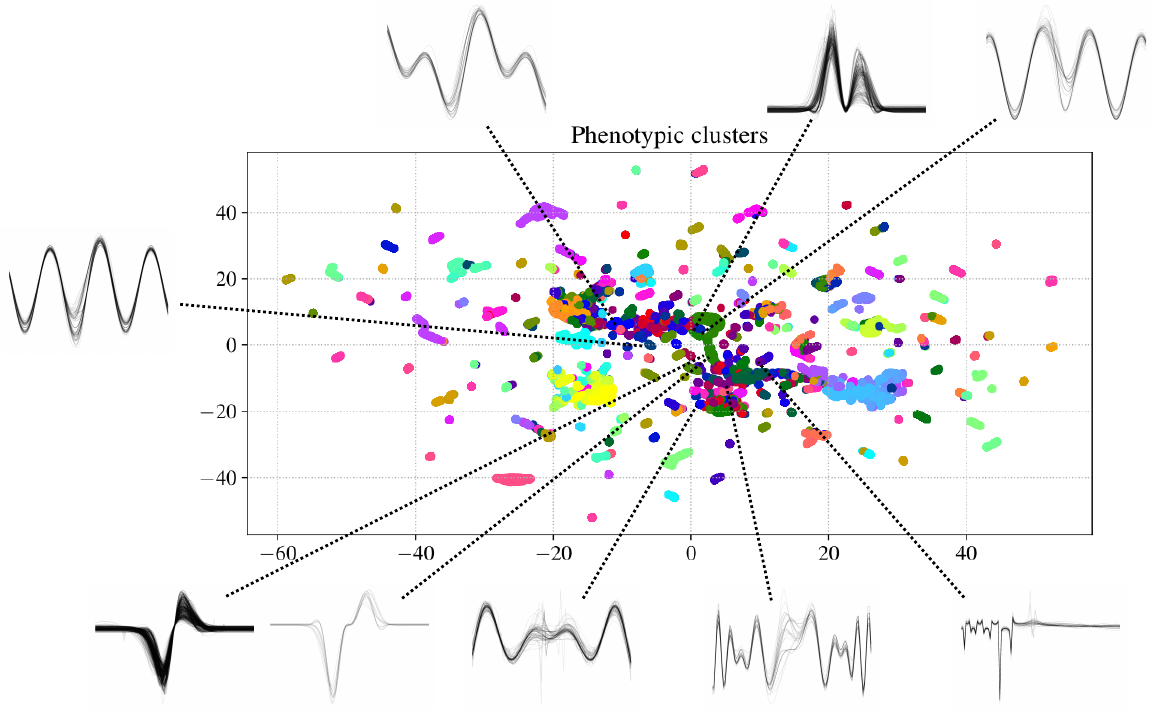}
  \caption{Visualization of the embedding and clustering result based on phenotypic similarity. The phenotypic mapping leads to several clearly defined clusters.}
  \label{fig:phenotypic-clustering}
\end{figure}

Figure \ref{fig:phenotypic-clustering} shows a result for the visualization and clustering based on phenotypic similarity where we have used LargeVis directly on all output vectors and using cosine-similarity. Each dot represents an expression. The color indicates to which cluster an expression has been assigned. The visualization clearly shows that several clusters of similar expressions can be identified in the search space.

As a post-processing step we prepared plots for all clusters which show the outputs of all expressions within the clusters. We found that the search space includes many rather complex functions and that several clusters of interesting and similar functions are identified. Some selected plots as well as the position of the cluster center on the mapped search space are shown in Figure \ref{fig:phenotypic-clustering}. It should be noted that the visualization does not show unique expressions which have been identified by HDBSCAN as outliers.  

\begin{figure}
  \includegraphics[width=\textwidth]{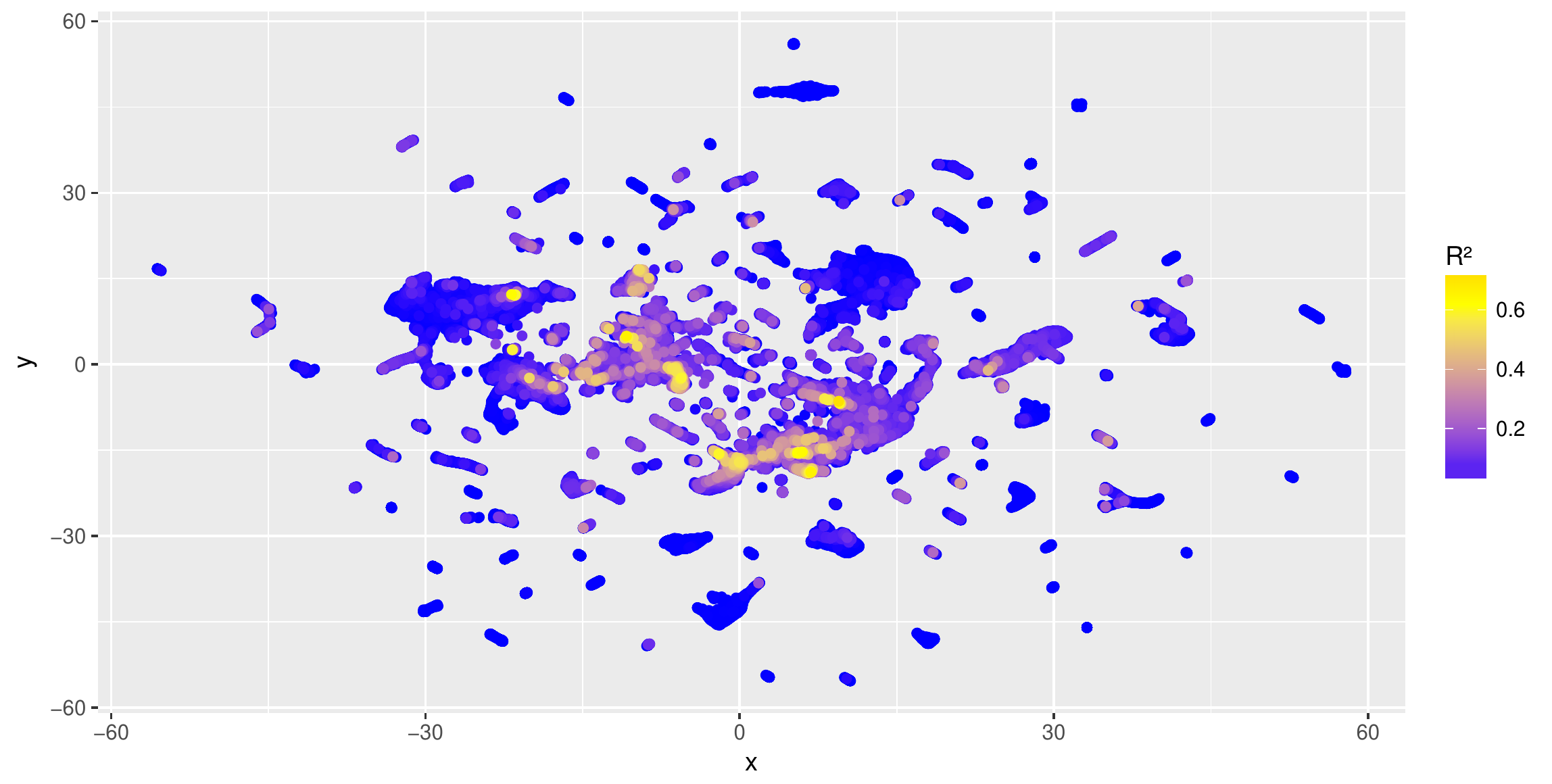}
  \caption{All expressions in the mapped search space. We use squared correlation with the Keijzer-4 function for the coloring.}
  \label{fig:phenotypic-map-keijzer4}
\end{figure}

Figure \ref{fig:phenotypic-map-keijzer4} again shows a phenotypic map. The difference to Figure \ref{fig:phenotypic-clustering} is that here all expressions are shown and we have used a coloring scheme based the similarity of expression outputs with the Keijzer-4 function (squared correlation $R^2$). The visualization clearly shows that only certain areas of the search space contain expressions which are similar to the target function. Notably, there are several areas which contain expressions with large $R^2$ values. There are at least two potential explanations for this. First, it could be an artifact of the approximation of t-SNE. Another reason could be the fact that we have used cosine similarity for the embedding and the $R^2$ value for the coloring scheme. With the cosine similarity measure, two vectors that are negatively correlated are dissimilar.        

\begin{figure}
  \includegraphics[width=\textwidth]{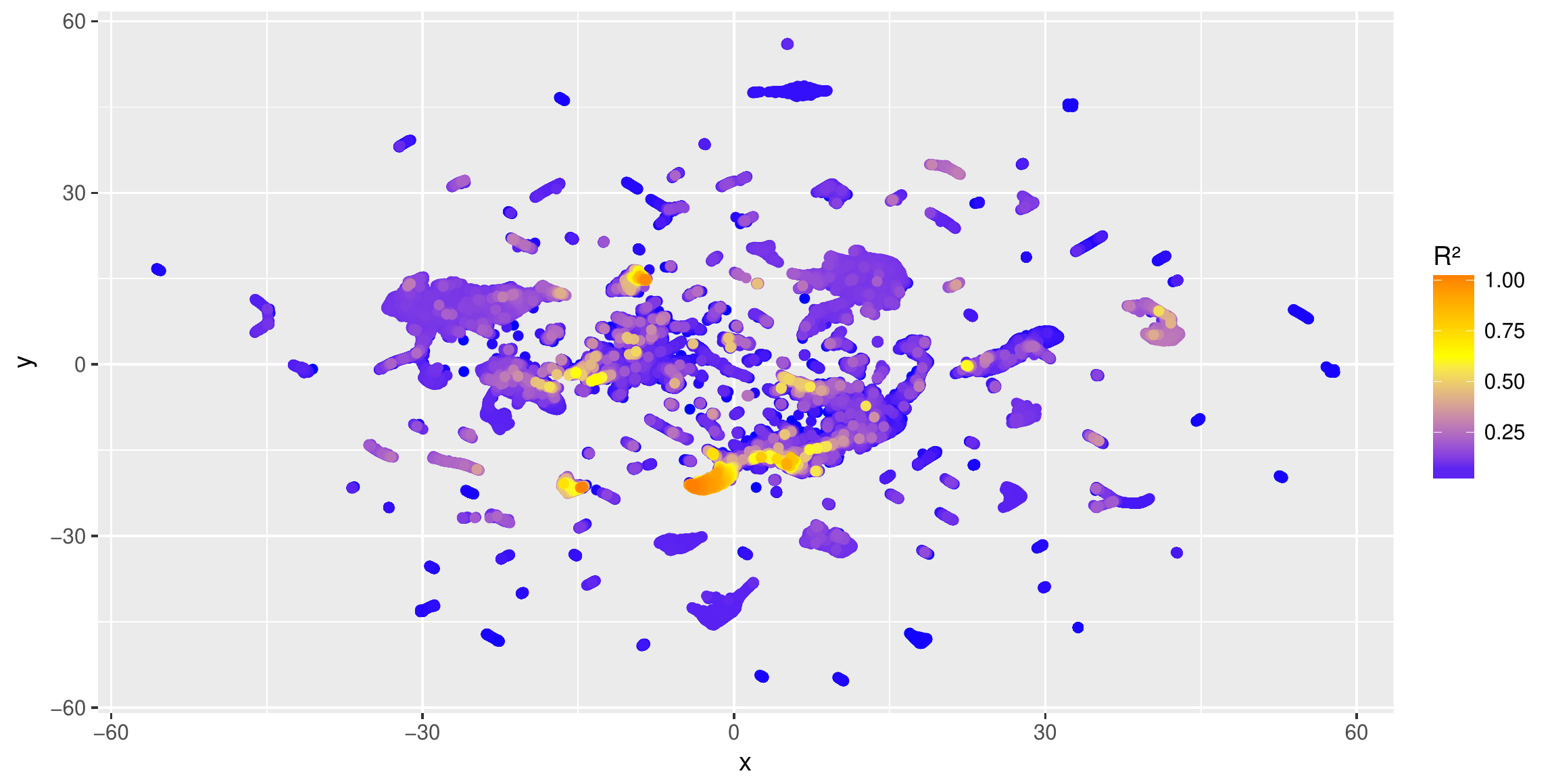}
  \caption{All expressions in the mapped search space colored based on the squared correlation with the Pagie-1d function.}
  \label{fig:phenotypic-map-pagie1d}
\end{figure}

Figure \ref{fig:phenotypic-map-pagie1d} shows the same visualization for a different function (Pagie-1d). Compared to Figure \ref{fig:phenotypic-map-keijzer4} other areas of the search space are highlighted.

The results produced for the phenotypic mapping are motivating for further research. At least for the two considered examples we should be able to use a hill-climbing algorithm on the mapped search space to find well-fitting expressions. The mapping of the search space must be pre-computed only once and can be reused for potentially any target function.    

\subsection{Genotypic mapping}
\begin{figure}
        \includegraphics[width=\textwidth]{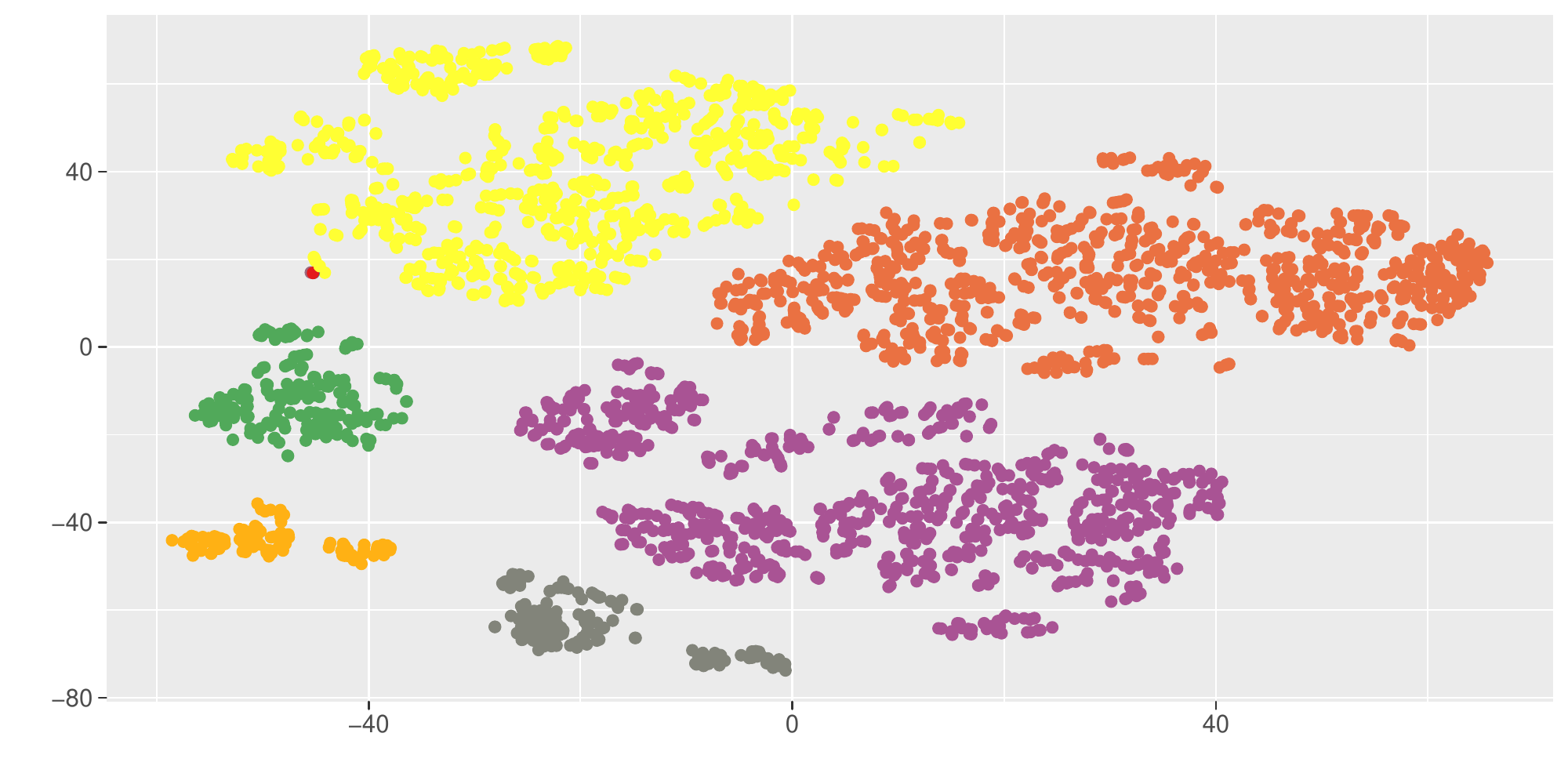}
        \includegraphics[width=\textwidth]{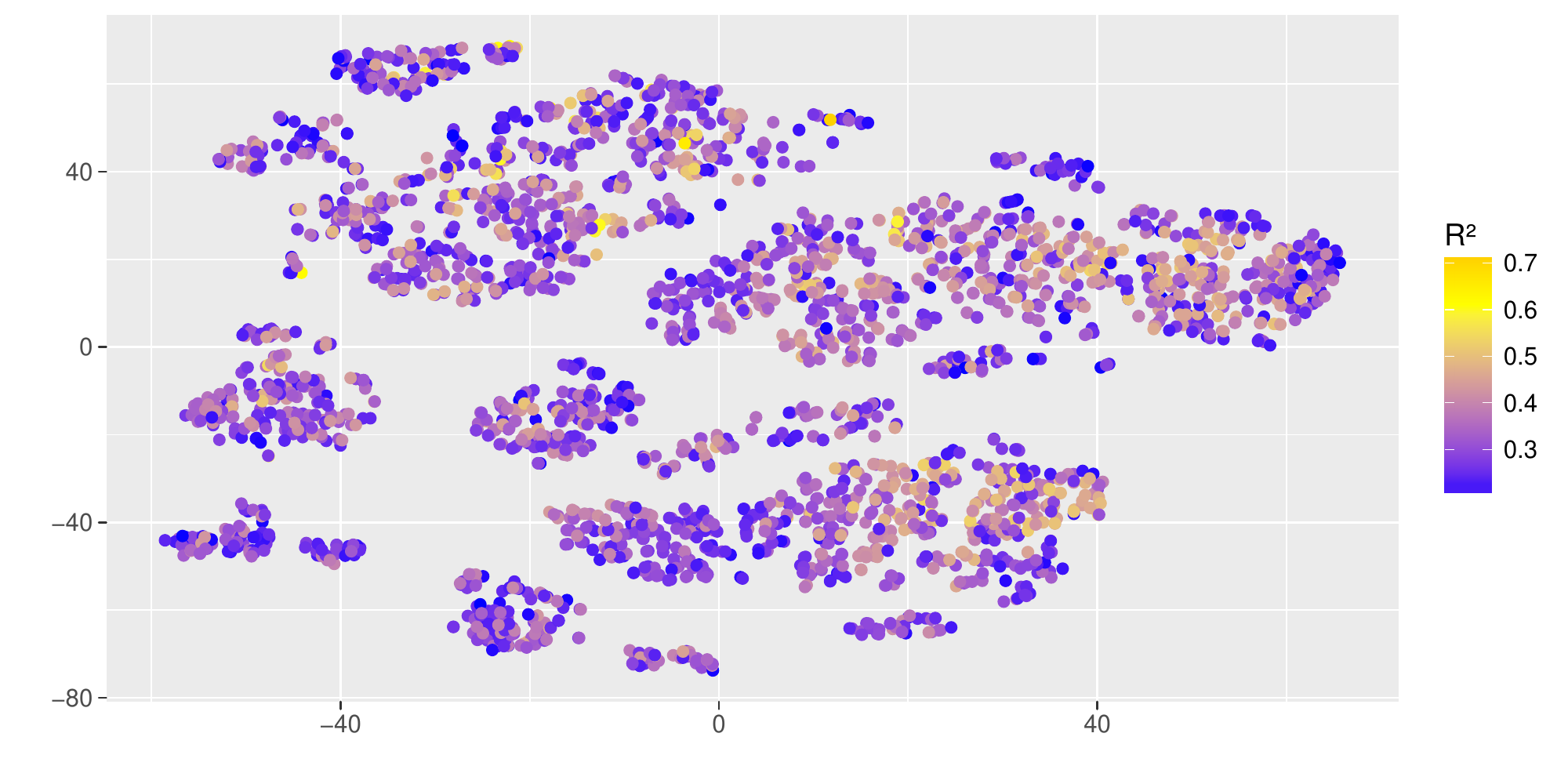}
        \caption{Results of HDBSCAN and t-SNE with the genotypic
          similarity (top: coloring based on clusters; bottom:
          coloring based on the $R^2$ value with the Keijzer-4
          function). The visualization of the two-dimensional
          mapping shows clusters of genotypically similar
          solutions. However, the genotypic clusters do not correlate
          strongly with qualities.}
        \label{fig:genotypic-clustering}
\end{figure}

Figure \ref{fig:genotypic-clustering} shows the results of t-SNE for dimensionality reduction and HDBSCAN for clustering when we use the genotypic similarity (see Section \ref{sec:genotypic-similarity}). We used HDBSCAN on the t-SNE mapped points. In comparison to the phenotypic mapping of the search space, the genotypic mapping does not produce such clearly defined clusters. In particular, in the lower sub-plot of Figure \ref{fig:genotypic-clustering} no strong correlation with the quality values is visible. 

\subsection{Cluster qualities for benchmark problems}
To check whether the phenotypic clustering of our search space is generalizable over multiple problem instances, we use the following five uni-variate
benchmark functions:
\begin{itemize}
\item Keijzer-4 \cite{keijzer2003improving}: $f(x) = x^3 e^{-x} \cos(x) \sin(x) (\sin(x)^2 \cos(x)-1)$ ~~for $x \in [0,10] $
\item Keijzer-9 \cite{keijzer2003improving}: $f(x) = \ln(x + \sqrt{x^2 + 1})$ ~~for $x \in [0,100]$
\item Pagie-1d \cite{pagie97}\footnote{We have used a uni-variate variant of the benchmark function described by Pagie and Hogeweg}: $f(x) = \frac{1}{1+x^{-4}}$ ~~for $x \in [-5,5]$
\item Nguyen-5 \cite{uy2011semantically}: $f(x) = \sin(x^2) \cos(x)- 1$ ~~for $x\in [-1,1]$
\item Nguyen-6 \cite{uy2011semantically}: $f(x) = \sin(x) + \sin(x+x^2)$ ~~for $x\in [-1,1]$
\end{itemize}

\begin{figure}
  \includegraphics[width=3.5cm]{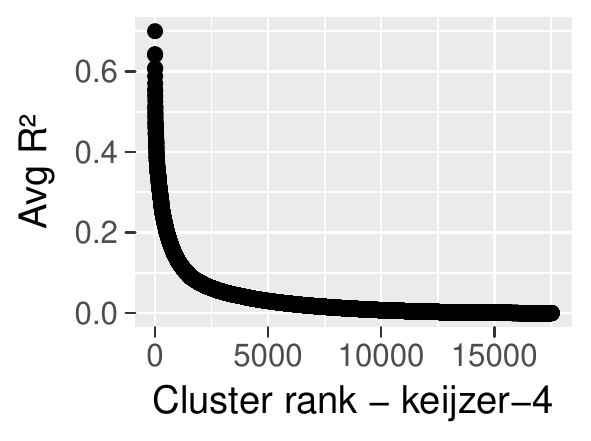}
  \includegraphics[width=3.5cm]{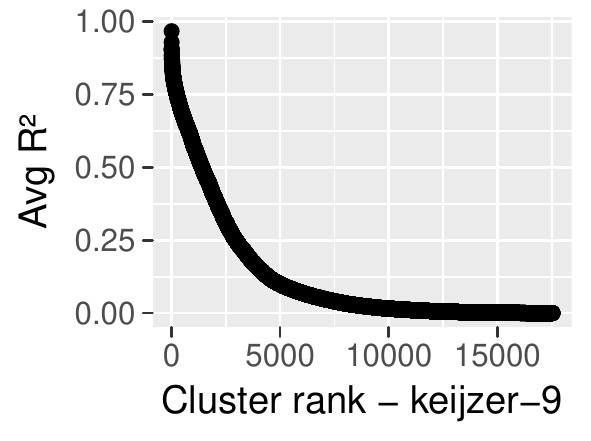}
  \includegraphics[width=3.5cm]{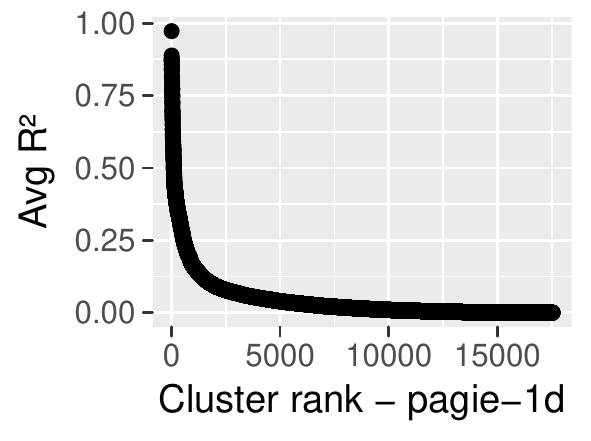}\\
  \includegraphics[width=3.5cm]{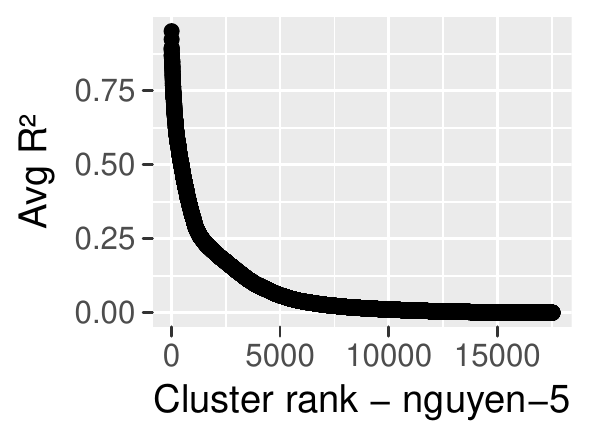}
  \includegraphics[width=3.5cm]{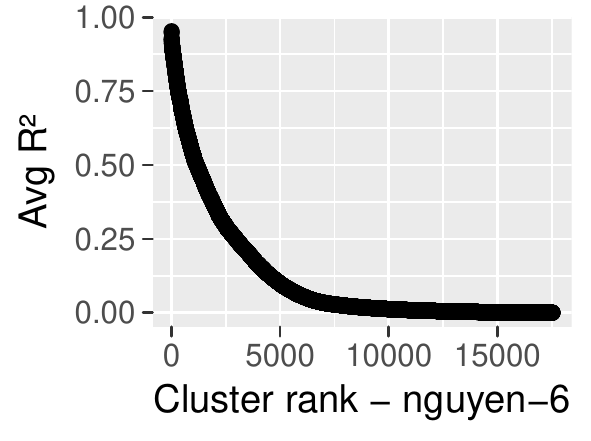}
        \caption{Ranking of clusters
          by average $R^2$ values of expressions within all
          clusters. For each of the benchmark functions there are
          clusters which contain well-fitting expressions. }
        \label{fig:avg-cluster-qualities}
\end{figure}

\begin{figure}

  Keijzer-4\\
  \includegraphics{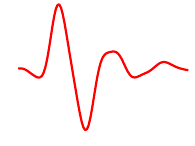}
  \includegraphics{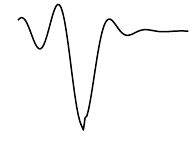}
  \includegraphics{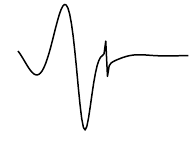}
  \includegraphics{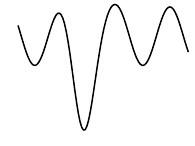}
  \includegraphics{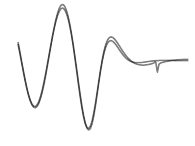}


  Keijzer-9\\
  \includegraphics{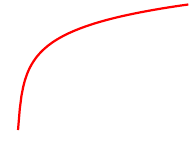}
  \includegraphics{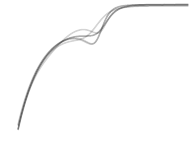}
  \includegraphics{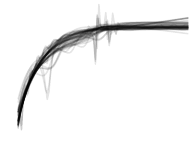}
  \includegraphics{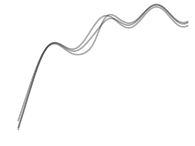}
  \includegraphics{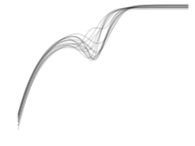}


  Pagie-1d\\
  \includegraphics{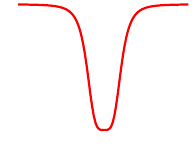}
  \includegraphics{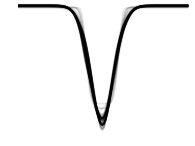}
  \includegraphics{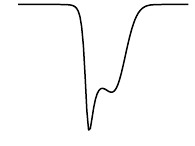}
  \includegraphics{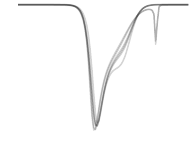}
  \includegraphics{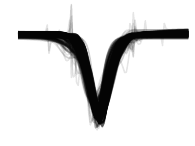}

  Nguyen-5\\
  \includegraphics{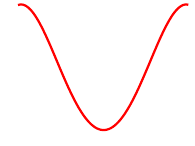}
  \includegraphics{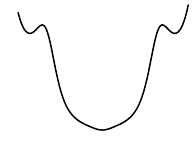}
  \includegraphics{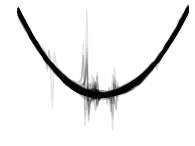}
  \includegraphics{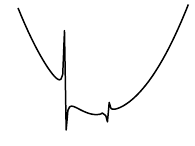}
  \includegraphics{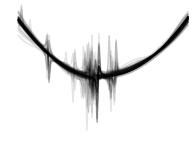}

  Nguyen-6\\
  \includegraphics{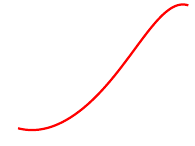}
  \includegraphics{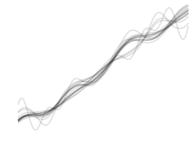}
  \includegraphics{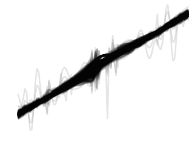}
  \includegraphics{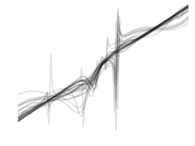}
  \includegraphics{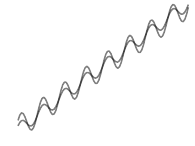}
  \caption{The five benchmark functions and the best four clusters
    with the highest average $R^2$ for each of the functions from left to right. The
    average $R^2$ values are: $0.70$ (Keijzer-4), $0.97$ (Keijzer-9), $0.97$ (Pagie-1d), $0.95$ (Nguyen-5), and $0.95$ (Nguyen-6).}
  \label{fig:best-clusters}
\end{figure}

We average the resulting $R^2$ values within each cluster and rank the
clusters by the average $R^2$ for each benchmark function
independently. Figure \ref{fig:avg-cluster-qualities} shows the
average $R^2$ values over cluster ranks for the five benchmark
functions. We find that for each of the benchmark functions a cluster
with well-fitting expressions is be found. Figure
\ref{fig:best-clusters} shows plots of all functions in the best four
clusters for each of the considered benchmark functions. The plots
show that for the Keijzer-4 and the Pagie-1d functions the search
space contains well-fitting expressions. However, for the other three
functions the best clusters do not fit the target function well. This is in contrast to the calculated average $R^2$ values for the clusters which are relatively high ($\geq 0.95$) for all functions except for Keijzer-4. Here
we want to stress again, that the search space clustering has been
calculated independently from the target functions.

\subsection{Mapping of GP solution candidates}
We use the 2d-mapping of the search space to analyze the search bias
of GP. For this we use a rather canonical GP implementation and map
each evaluated solution candidate to the search space by identifying
the closest representative expression in the enumerated search
space. Each expression in the search space is assigned to a cluster. Therefore, we can determine which clusters of the search space are visited by GP. We have again used the Keijzer-4 function to
demonstrate the concept.

\begin{figure}
  \includegraphics[width=0.5\textwidth]{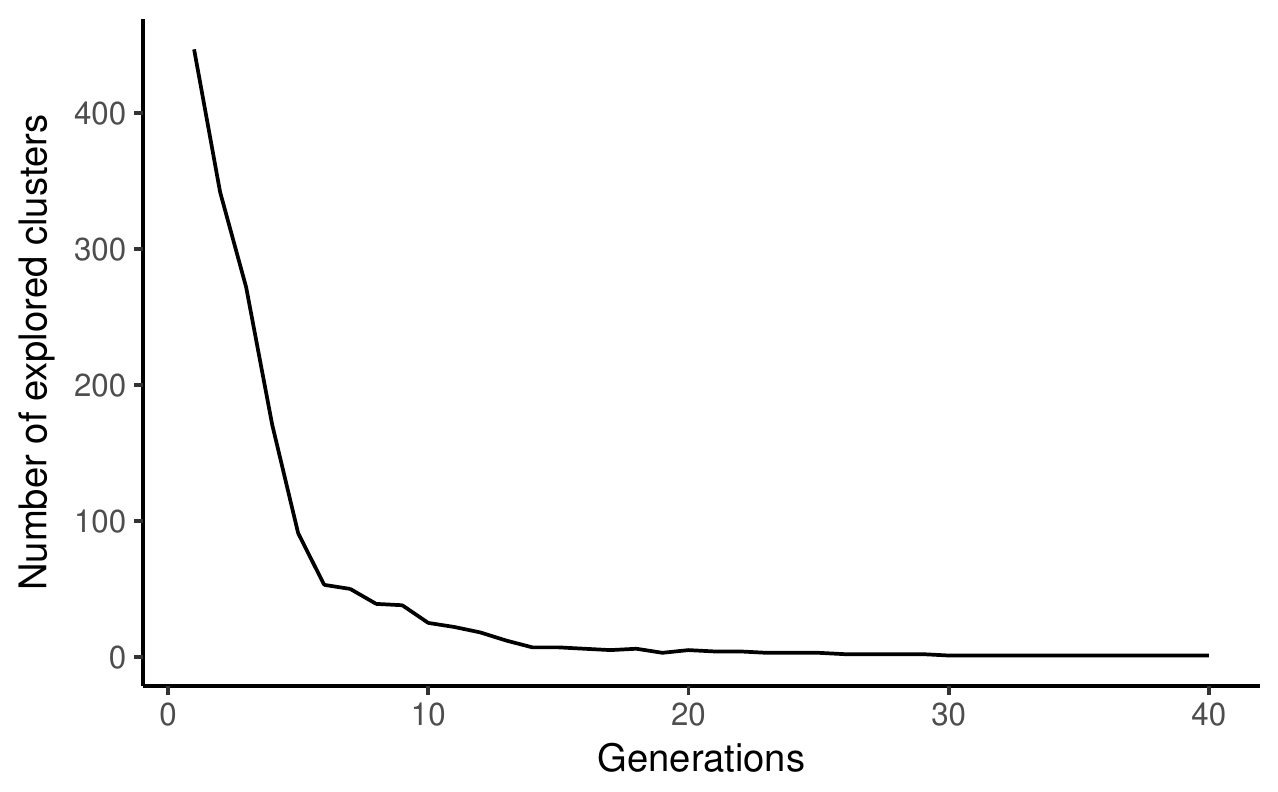}
  \includegraphics[width=0.5\textwidth]{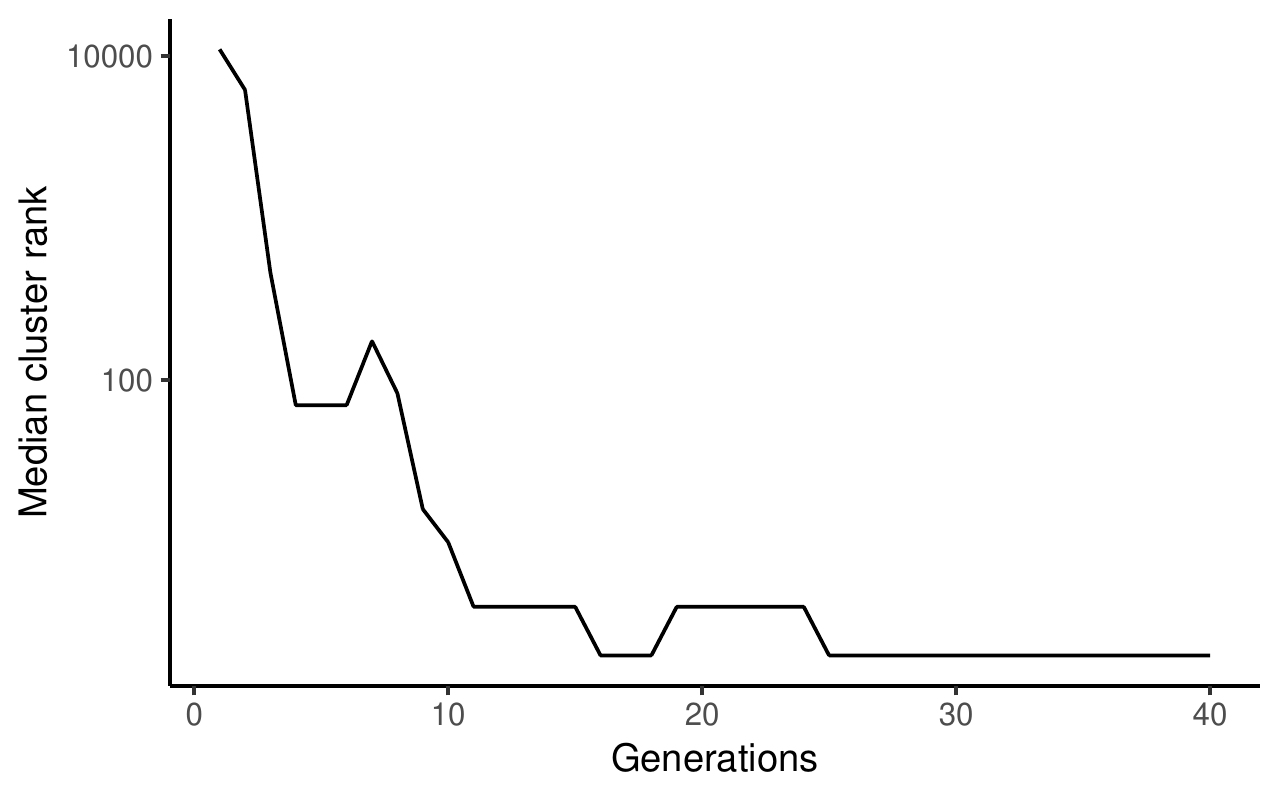}
  \caption{The number of clusters that are visited by GP as well as
    the median rank (quality) of clusters over GP generations.}
  \label{fig:gp-explored-clusters}
\end{figure}

Figure \ref{fig:gp-explored-clusters} shows the results of the analysis. In the left sub-plot the number of different clusters visited by GP are shown over generations; on the right hand side the median cluster rank (ordered by cluster quality) is shown. This clearly shows that in the beginning GP visits many different solution candidates and later concentrates on only a view high quality clusters. 

\begin{figure}
  \includegraphics[width=0.3\textwidth]{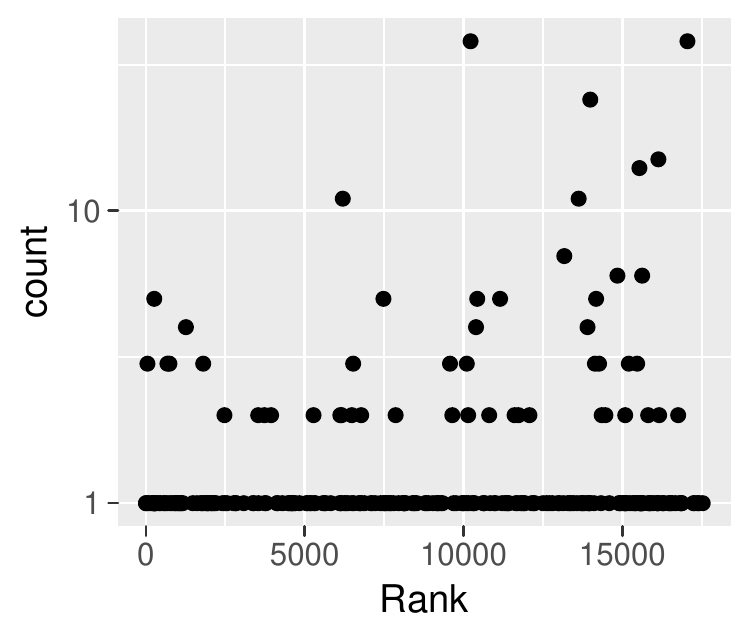}
  \includegraphics[width=0.3\textwidth]{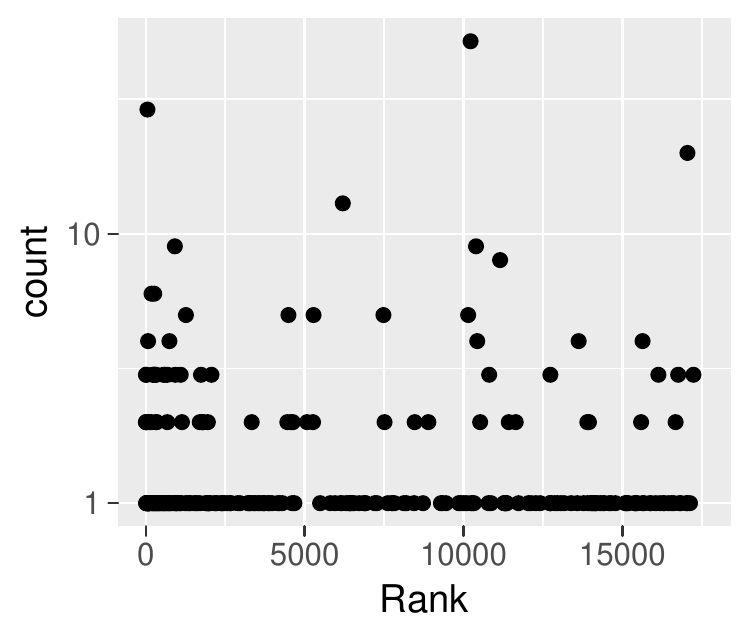}
  \includegraphics[width=0.3\textwidth]{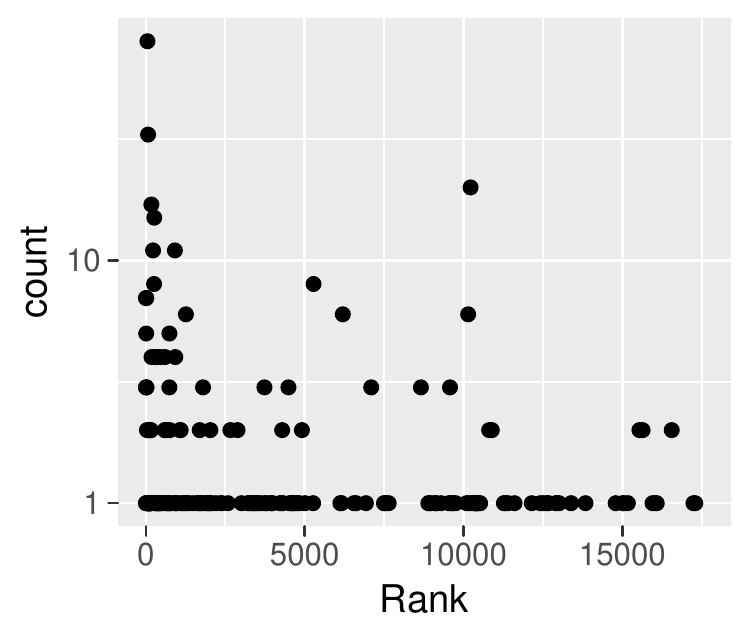}\\
  \includegraphics[width=0.3\textwidth]{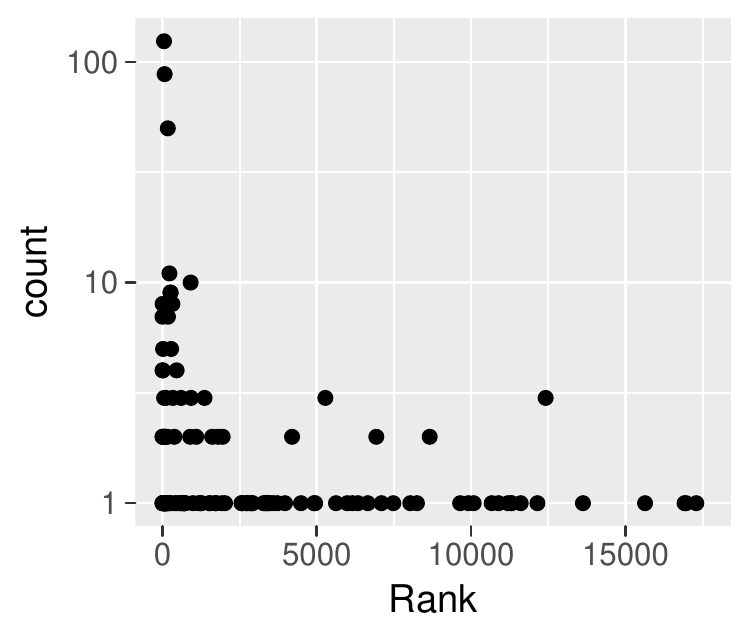}
  \includegraphics[width=0.3\textwidth]{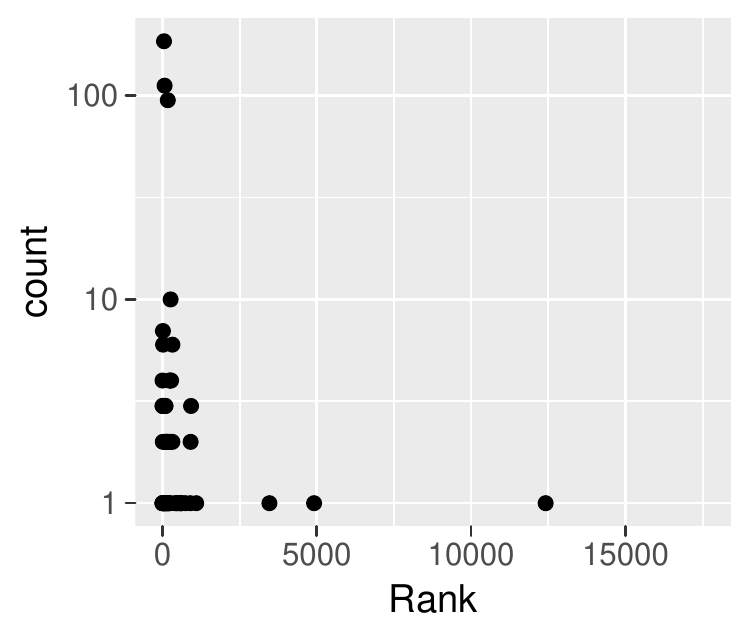}
  \includegraphics[width=0.3\textwidth]{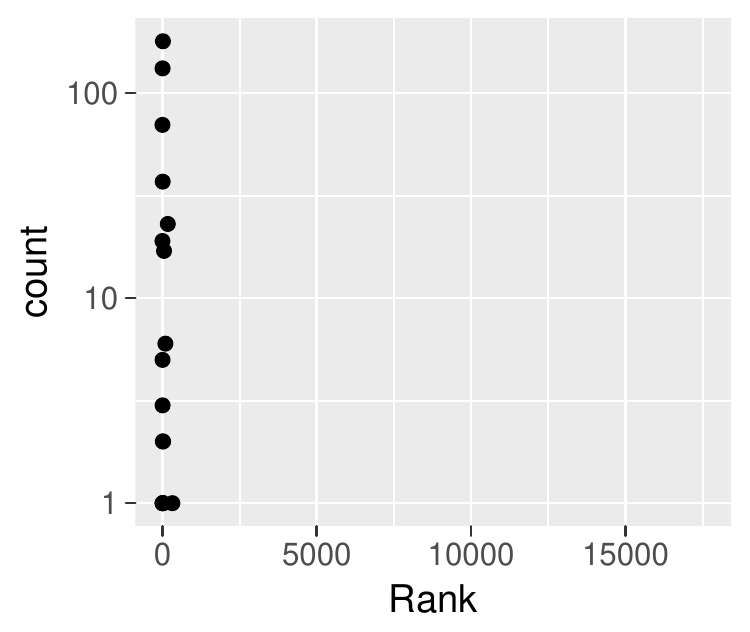}\\

  \includegraphics[width=0.3\textwidth]{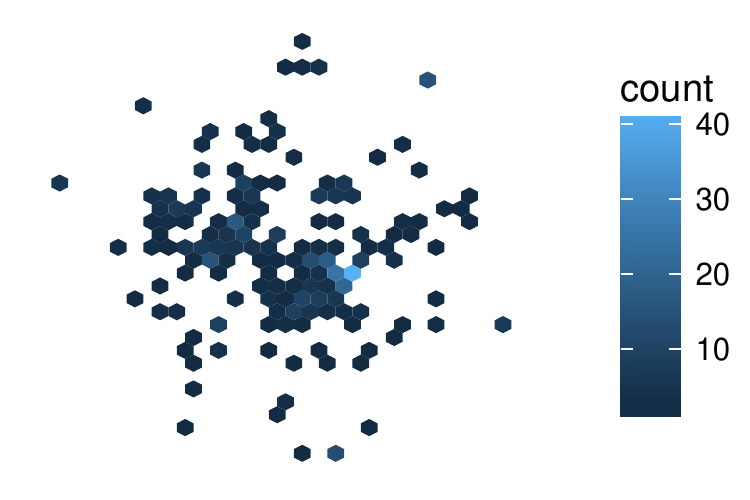}
  \includegraphics[width=0.3\textwidth]{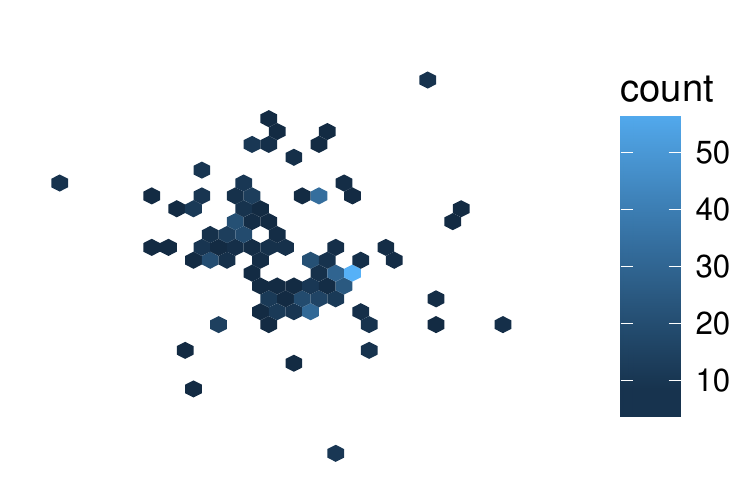}
  \includegraphics[width=0.3\textwidth]{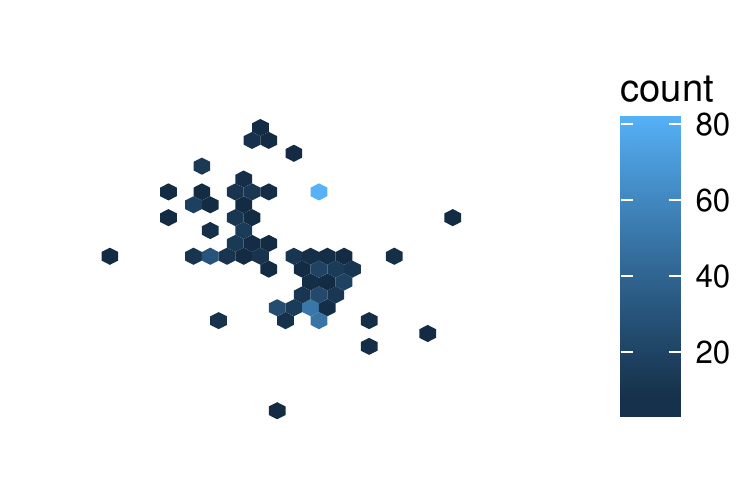}\\
  \includegraphics[width=0.3\textwidth]{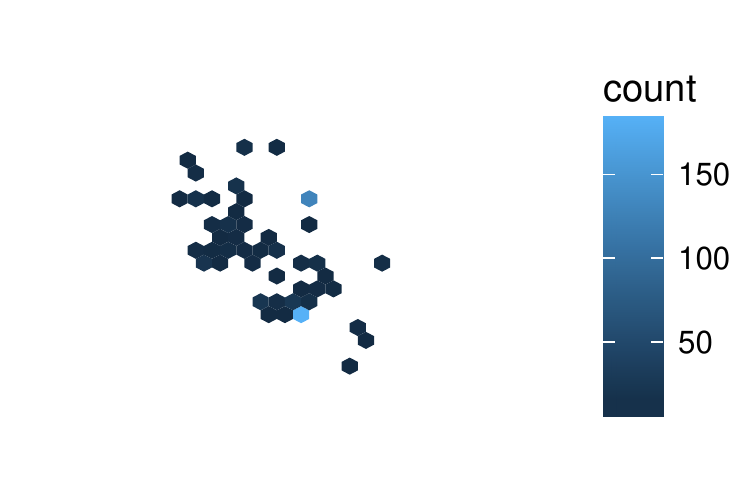}
  \includegraphics[width=0.3\textwidth]{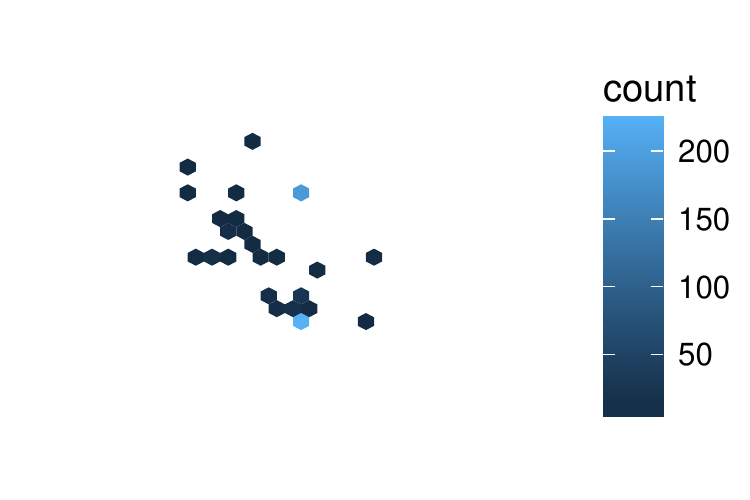}
  \includegraphics[width=0.3\textwidth]{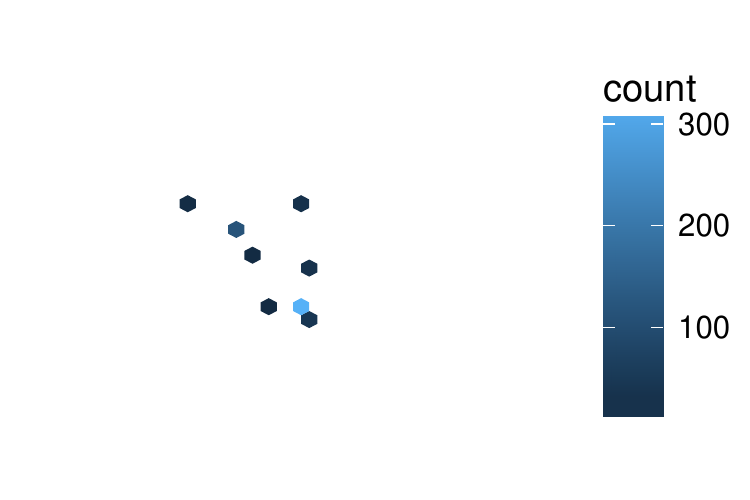}
    
  \caption{More detailed visualization of the search space explored by a GP run. The plots in the first and second rows show the visitation frequency for each cluster in generations $1, 2, 3, 4, 5, 10$. Clusters with lower ranks are high quality clusters. The plots in the third and fourth row show the visitation frequency in the two-dimensional phenotypic map.  Within the first 10 generations GP converges to a focused area of the search space.}
  \label{fig:gp-explored-clusters-detail}
\end{figure}

In Figure \ref{fig:gp-explored-clusters-detail} we show the distribution of solution candidates visited by GP in more detail for the first few generations (1,2,3,4,5,and 10). It is clearly visible that within the first ten generations GP explores almost each of the 16000 clusters (population size=500 and PTC2 tree creator \cite{Luke2000}) and quickly finds the clusters with highest quality. 

\section{Discussion}
\label{sec:discussion}
 Our analysis so far has some limitations that merit a more detailed
 discussion.
 \begin{itemize}
 \item We have only looked at uni-variate models.
 \item The grammar is very restricted.
 \item Even with the limit of seven variable references, the
   computational effort is already rather high. When increasing the
   length limits the computational effort quickly becomes too big.
 \item The phenotypic clustering depends on the range of input values
   that are used when evaluating the model.
 \item We have completely ignored the effect of numeric constants in
   models, i.e. we have not used numeric constants.
 \end{itemize}

 In this work we have considered a search space of approximately
 160000 semantically different expressions. However, in Figure
 \ref{fig:best-clusters} we see that the search space does not contain
 well-fitting expressions for all of the considered target
 functions. For a more in-depth analysis the size of
 the search space should be extended even for uni-variate problems.

 We also found that the grammar we used produced many discontinuous
 functions (e.g. because of division by zero or extreme arguments to
 $\exp(x)$). If we assume that we are not interested in such functions
 then the search space could potentially be reduced massively by
 removing expressions leading to discontinuous functions.
 
 If we extend the analysis to include multi-variate models, the size
 of the search space would increase significantly even if we use the
 same size restrictions. This is simply a consequence of the fact that
 there are more different models that can be expressed. Based on
 preliminary experiments, even with two or three independent variables
 it is possible to enumerate the search space with the same size
 restrictions as we used above. For more than three variables it would
 be necessary to use even smaller size limits.

 We also need to consider that the variety of function outputs becomes
 much larger with increased dimensionality which could lead to a
 larger set of clusters.

 We hypothesize that for practical problems it is usually sufficient
 to be able to represent two- or three-way interactions of variables
 as separable terms can be modeled independently. However, we cannot
 expect this in general.

 Regarding the complexity of the grammar, we have purposefully limited
 the number of alternatives to be able to enumerate the full search
 space. It would however be rather easy to add more functions (e.g. a
 power or root function) as long as the complexity of the argument to
 the added functions is limited similarly as we have limited arguments for $\log(x)$.

 A different approach that could potentially be worthwhile is to
 calculate the phenotypic similarity in the frequency-domain of the
 function.

\section{Conclusion}
\label{sec:conclusion}
This contribution aims to analyze the search behavior of GP in a space of hypotheses that is visualized as a 2D mapping of the search space. The idea of this approach is to enumerate the complete search space in the off-line phase independently of the regression problem to be solved. In order to do so we had to define some restrictions like the consideration of only uni-variate functions under restricting grammar assumptions. The still huge search space can be further restricted by filtering unique expressions by hashing. Once having establishes a mapping of the so defined search space by hierarchical clustering this off-line generated setup can be used to monitor the search behavior of different algorithms can be monitored and analyzed as demonstrated in this chapter for a relatively canonical GP implementation for the genotypic as well as for the phenotypic search space.

The ideas presented in this chapter have to be considered as very first findings of a novel approach having in mind that the restrictions on the search space delimit the generality of claimed findings. In order to achieve a deeper and more universal understanding it will be necessary to extend the approach to multi-variate models and less restrictions on the grammar of the model. Also it will be interesting to analyze the search behavior of different flavors of GP and other hypothesis search techniques.   

Furthermore, the massive set of off-line generated models could be used in order to filter out an initial population for a GP run as soon as the regression problem to be tackled is available. Similar to what is done in the ANN community recently with pre-trained neural networks we could establish somehow pre-evolved initial populations for evolutionary search: In this way we could for example filter a genotypically diverse subset of model with promising correlation problems for a concrete problem in order to establish a somehow pre-evolved initial population for a GP run which is aimed to converge a lot faster in the on-line stage.

\begin{acknowledgement}
  The authors thank the participants of the Genetic Programming in Theory and Practice (GPTP XVI) workshop for their valuable feedback and ideas which helped to improve the work described in this chapter.
  
  The authors gratefully acknowledge support by the Christian Doppler
  Research Association and the Federal Ministry for Digital and
  Economic Affairs within the \emph{Josef Ressel Center for Symbolic
    Regression}.
\end{acknowledgement}

\bibliographystyle{spmpsci}
\bibliography{bibliography}
\end{document}

%% file: abstract.tex
In this chapter we take a closer look at the distribution of symbolic regression models generated by genetic programming in the search space.
The motivation for this work is to improve the search for well-fitting symbolic regression models by using information about the similarity of models that can be precomputed independently from the target function.
For our analysis, we use a restricted grammar for uni-variate symbolic regression models and generate all possible models up to a fixed length limit. We identify unique models and cluster them based on phenotypic as well as genotypic similarity.
We find that phenotypic similarity leads to well-defined clusters while genotypic similarity does not produce a clear clustering. By  mapping solution candidates visited by GP to the enumerated search space we find that GP initially explores the whole search space and later converges to the subspace of highest quality expressions in a run for a simple benchmark problem. 